\documentclass{article}

\usepackage{microtype}
\usepackage{graphicx}
\usepackage{subfigure}
\usepackage{booktabs} 

\usepackage{amsthm}
\usepackage{amsmath}
\usepackage{amssymb}

\renewcommand{\Re}{\mathbb{R}}
\renewcommand{\S}{Section }

\theoremstyle{definition}

\usepackage{siunitx}
\usepackage{bm}
\usepackage{bbm}

\newcommand{\demopageurl}{https://google.github.io/tacotron/publications/end_to_end_prosody_transfer}
\newcommand{\demopagelink}[1]{\href{\demopageurl}{#1}}

\usepackage{hyperref}

\usepackage[accepted]{icml2018}

\icmltitlerunning{Towards End-to-End Prosody Transfer for Expressive Speech Synthesis with Tacotron}

\begin{document}

\twocolumn[
\icmltitle{Towards End-to-End Prosody Transfer \\ for Expressive Speech Synthesis with Tacotron}




\begin{icmlauthorlist}
\icmlauthor{RJ Skerry-Ryan}{google}
\icmlauthor{Eric Battenberg}{google}
\icmlauthor{Ying Xiao}{google}
\icmlauthor{Yuxuan Wang}{google}
\icmlauthor{Daisy Stanton}{google}
\icmlauthor{Joel Shor}{google}
\icmlauthor{Ron J. Weiss}{google}
\icmlauthor{Rob Clark}{google}
\icmlauthor{Rif A. Saurous}{google}
\end{icmlauthorlist}

\icmlaffiliation{google}{Google, Inc.}

\icmlcorrespondingauthor{RJ Skerry-Ryan}{rjryan@google.com}

\icmlkeywords{Machine Learning, ICML}

\vskip 0.3in
]



\printAffiliationsAndNotice{}  

\begin{abstract}
We present an extension to the Tacotron speech synthesis architecture that learns a latent embedding space of prosody, derived from a reference acoustic representation containing the desired prosody.
We show that conditioning Tacotron on this learned embedding space results in synthesized audio that matches the prosody of the reference signal with fine time detail even when the reference and synthesis speakers are different.  Additionally, we show that a reference prosody embedding can be used to synthesize text that is different from that of the reference utterance. 
We define several quantitative and subjective metrics for evaluating prosody transfer, and report results with accompanying audio samples from single-speaker and 44-speaker Tacotron models on a prosody transfer task.
\end{abstract}

\newtheorem*{definition*}{Definition}\theoremstyle{definition}

\section{Introduction}
\label{sec:introduction}

In order to produce realistic speech, a text-to-speech (TTS) system must implicitly or explicitly impute many factors that are not given in a simple text input. 
Such factors include the intonation, stress, rhythm and style of the speech, and are collectively referred to as \textit{prosody}.

Speech synthesis via text-to-speech is a challenging underdetermined problem, since the meaning expressed by an utterance is inherently underspecified by the text. For example, the simple statement ``The cat sat on the mat.'' can be spoken many different ways. If the statement is the answer to the question ``Where did the cat sit?'' the speaker might stress the word ``mat'' to indicate that it is the answer to the question. To express uncertainty in their knowledge, the speaker may decide to intone the response with a rising pitch. The question, ``Would you like an apple or an orange?'' can also be spoken in multiple ways, indicating information about the set of objects that exist. If there are only two possible options, the intonation of the final option (``orange'') will have a declining pitch. If there are a variety of options of which apple and orange are just two examples, both options are typically intoned with a rising pitch. The intonation of these sentences carries meaning about the environment or context of the question which is unspecified by the text, and in general, there are any number of such nuances present in speech that convey information beyond the textual content.

In order to avoid the challenging problem of schematizing and labeling prosody, we seek methods of modeling prosody that do not require explicit annotations, and present an architecture for learning a latent prosody representation by extracting it from the ground truth speech audio. Accordingly, we use a ``subtractive'' definition of prosody: 
\begin{definition*} \label{def:prosody}
\textit{Prosody is the variation in speech signals that remains after accounting for variation due to phonetics, speaker identity, and channel effects (i.e. the recording environment}).
\end{definition*}
This view of prosody is compatible with interpretations of prosody from previous works \cite{wagner2010experimental,ladd2008intonational}. 

One natural problem that arises from this formulation is \emph{sampling} -- that is, the challenge of generating diverse and interesting prosody and output speech even for identical phonetics, speaker identities and channel effects. In this paper, we tackle the more basic problem of \emph{constructing} a space that represents prosody. We propose one possible construction of a prosody latent space, and show that we capture meaningful variation in speech by demonstrating transfer in this space (i.e., using a latent representation to make one utterance sound like another): this roughly corresponds to a ``say it like this'' task.

The recently proposed Tacotron speech synthesis system \cite{wang2017} computes its output directly from graphemes or phonemes, and its prosody model is implicit, learned from the statistics of the training data alone. It learns, for example, that an English sentence ending in a question mark likely has a rising pitch if the question has a yes-or-no answer.
In this work, we augment Tacotron with explicit prosody controls.
We accomplish this by learning an encoder architecture that computes a low-dimensional embedding from a speech signal, where the embedding provides information not provided by the text and speaker identity.
Through careful experiments, we demonstrate that this prosody embedding can be used to reproduce the desired prosody using Tacotron.

The immediate implication of this acoustic encoder architecture and prosody latent space is that we can control the behavior of a TTS system using a different voice than the one used in training. The resulting embedding is fixed-length and often smaller than the transcript, so it can be easily stored alongside the text for use in a production system. The longer-term implications are that we can build models that predict prosody embeddings from non-acoustic context, such as prosody labels or conversation state.

Our main contribution is an encoder architecture that extracts a fixed-length learned representation of prosody from acoustic input; we demonstrate that this encoder allows us to \emph{transfer} prosody between utterances in an almost speaker-independent fashion. To evaluate performance in this prosody transfer task, we propose a number of quantitative and qualitative metrics. Additionally, we strongly encourage the reader to listen to the audio samples on our \demopagelink{demo page}.
\section{Related Work}
\label{related-work}

Prosody and speaking style modeling have been studied since the era of HMM-based TTS research.
For example, \cite{eyben2012unsupervised} proposes a system that first clusters the training set, and then performs HMM-based cluster-adaptive training. 
\cite{nose2007style} proposes estimating the transformation matrix for a set of predefined style vectors.

Numerous works have explored annotation schemes for diagramming and automatic labeling of prosody: ToBi \cite{silverman1992tobi}, AuToBI \cite{rosenberg2010autobi}, Tilt \cite{taylor1998tilt}, INTSINT \cite{hirst2001automatic}, SLAM \cite{obin2014slam} all describe methods for the annotation and automatic extraction of labels or annotations that correlate with prosodic phenomena. The challenges of annotation often require domain experts,however, and inter-rater annotations can differ substantially \cite{wightman2002tobi}.

Few works propose the use of acoustic reference signals to control the prosody of a text-to-speech model. \cite{tesser2013experiments} proposes the use of ``signal driven'' features to predict symbolic prosody representations, using AuToBI labels to improve HMM-based synthesis. \cite{coile1994protran} propose ``prosody transplantation'' via a system called PROTRAN for recording a low-bit-rate ``enriched phonetic transcription'' that can be used in conjunction with desired text to reproduce the prosody of an original recording. Note that the same product needs described in \cite{coile1994protran} motivate the development of this paper.

Prosody transfer is related to the task of voice conversion (also called style transfer in the audio context). To perform voice conversion, a model must synthesize an utterance, given only the acoustic signal of that utterance in a different speaker's voice ~\cite{Zhizheng13,Nakashika2016,Tomi2017VC,van2017neural,chorowski2017using}. An approach similar to ours can be found in \cite{wang2018}, where a more complicated autoencoder is used to learn some elements of style in an unsupervised fashion.
\section{Model Architecture}
\label{sec:architecture}

\begin{figure*}[t]
\vskip 0.2in
\begin{center}
\centerline{\includegraphics[scale=0.6]{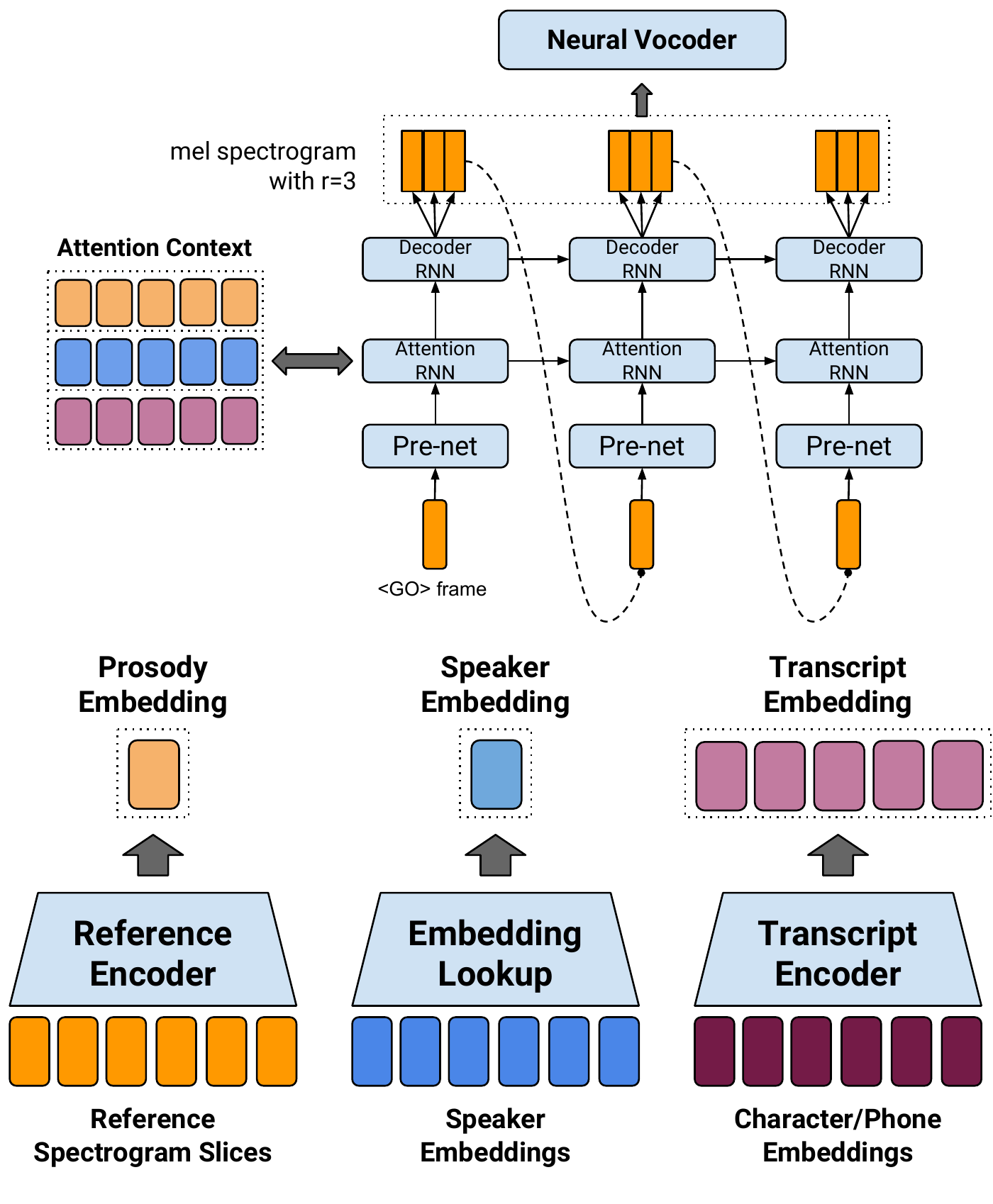}}
\caption{The full Tacotron architecture for prosody control. The autoregressive decoder is conditioned on the result of the reference encoder, transcript encoder, and speaker embedding via an attention module.} 
\label{fig:tacotron-conditioning}
\end{center}
\vskip -0.2in
\end{figure*}

\begin{figure}[t]
\vskip 0.2in
\begin{center}
\centerline{\includegraphics[height=6cm]{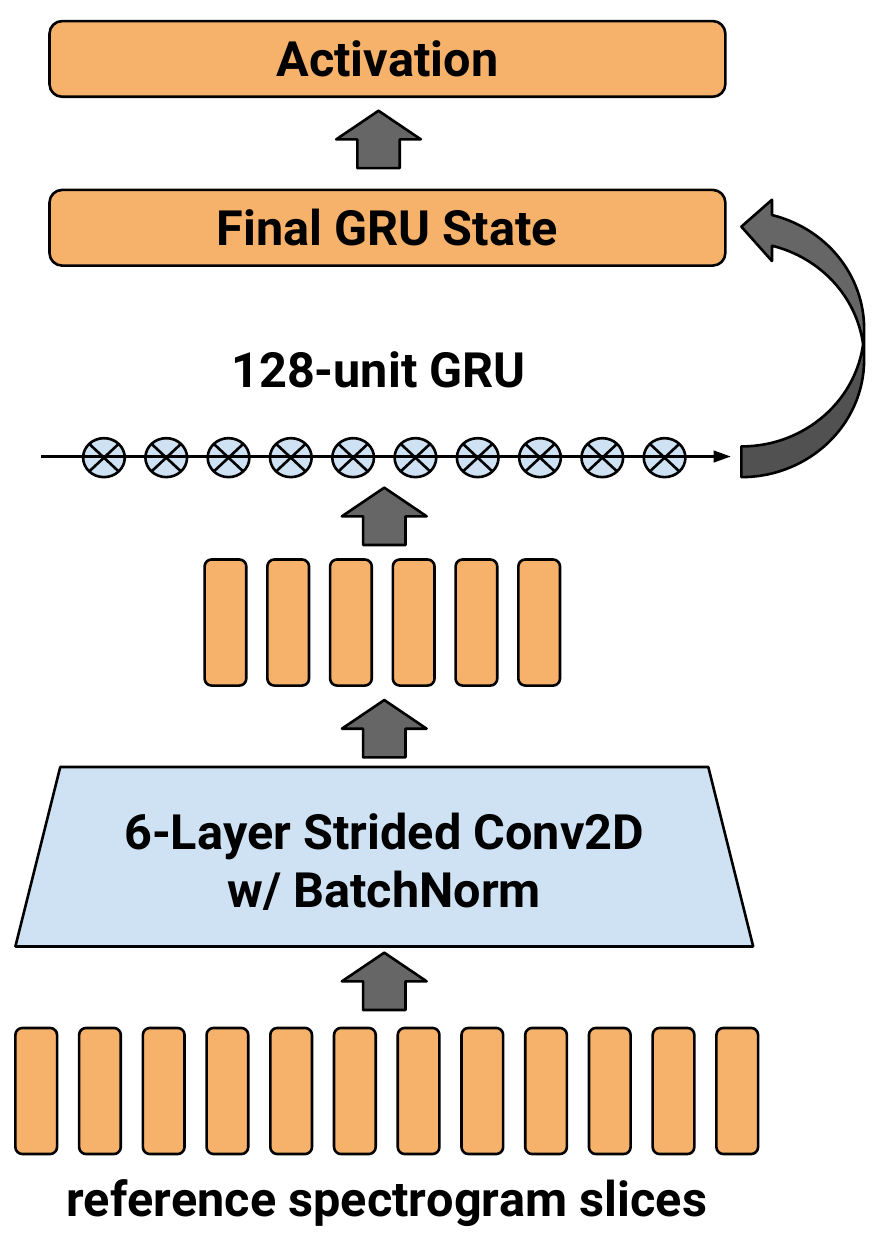}}
\caption{The prosody reference encoder module. A 6-layer stack of 2D convolutions with batch normalization, followed by ``recurrent pooling'' to summarize the variable length sequence, followed by an optional fully connected layer and activation.}
\label{fig:reference-encoder}
\end{center}
\vskip -0.2in
\end{figure}

\begin{figure}[t]
\vskip 0.2in
\begin{center}
\centerline{\includegraphics[height=6cm]{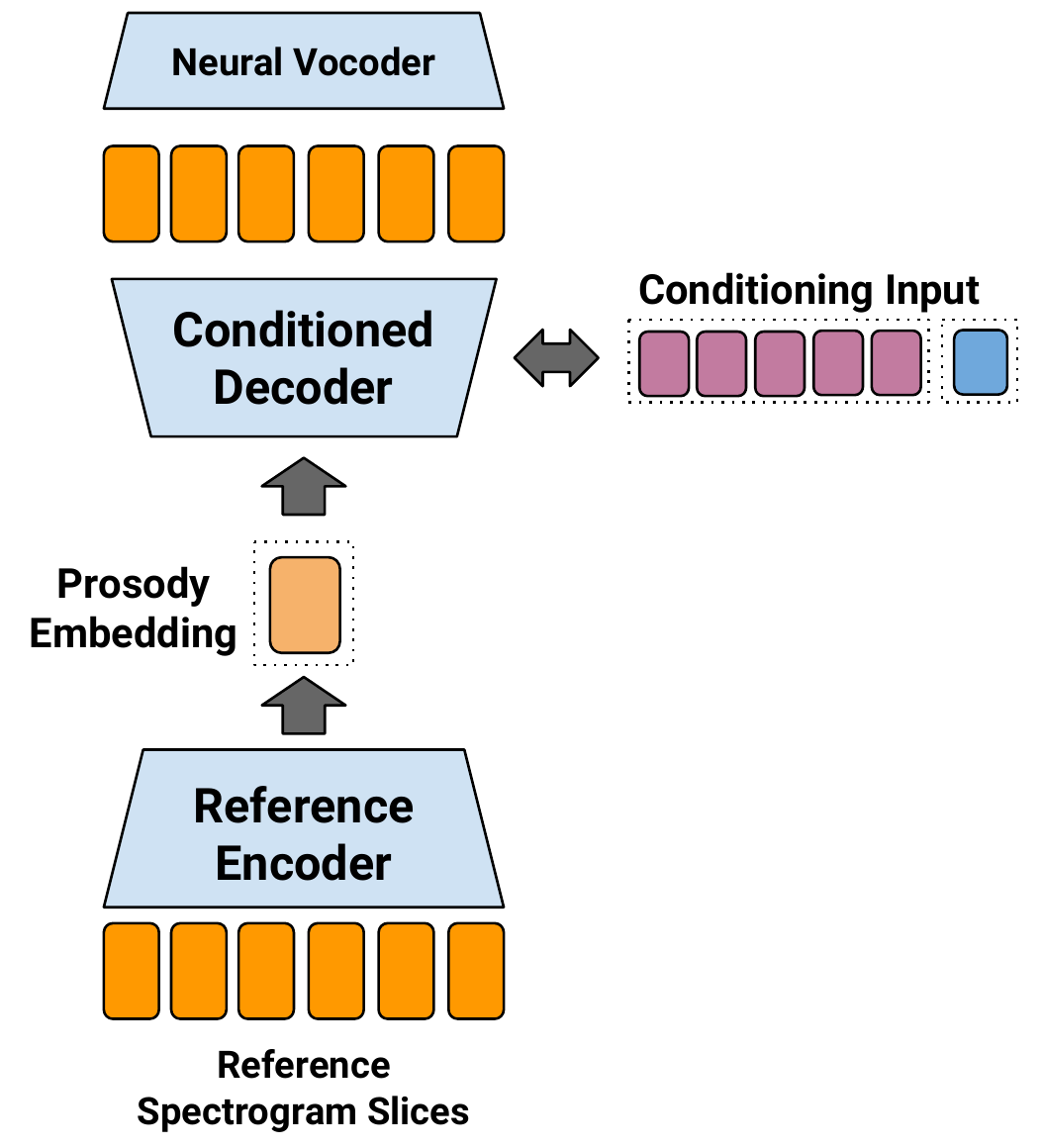}}
\caption{An interpretation of the Tacotron architecture for prosody control from Figure \ref{fig:tacotron-conditioning} as an RNN encoder-decoder with speaker and phonetic conditioning input.}
\label{fig:tacotron-autoencoder}
\end{center}
\vskip -0.2in
\end{figure}

Our model is based on Tacotron \cite{wang2017}, a recently proposed state-of-the-art end-to-end speech synthesis model that predicts mel spectrograms directly from grapheme or phoneme sequences. The predicted mel spectrograms can either be synthesized directly to the time-domain via a WaveNet vocoder \cite{shen2017}, or by first learning a linear spectrogram prediction network, and then applying Griffin-Lim spectrogram inversion \cite{griffin1984signal}.

In this work, we use the original encoder and decoder architecture from \cite{wang2017}, not the simplified architecture proposed by \cite{shen2017}. Additionally, we exclusively use phoneme inputs produced by a text normalization front-end and lexicon, as we are specifically interested in addressing prosody, not the model's ability to learn pronunciation from graphemes. Finally, instead of the Bahdanau attention used in \cite{wang2017}, we use the GMM attention of \cite{graves2013generating} which we find improves generalization to long utterances.

The audio samples included on our \demopagelink{demo page} were produced with a WaveNet vocoder \cite{shen2017}; however, the original linear-spectrogram prediction network followed by Griffin-Lim spectrogram inversion from \cite{wang2017} works equally well for prosody transfer. In practice, we find the choice of neural vocoder only impacts audio fidelity and has no impact on the system's resulting prosody.

\subsection{Multi-speaker Tacotron}
\label{sec:multi-speaker}

Tacotron as proposed in \cite{wang2017} does not include explicit modeling of speaker identity; however, due to the flexibility of all neural sequence-to-sequence models, learning multi-speaker models via conditioning on speaker identity is straightforward. We follow a scheme similar to \cite{arik2017deep} to model multiple speakers.

The Tacotron architecture conditions an auto-regressive decoder on an $L_T \times d_T$-dimensional representation of the phoneme or grapheme sequence produced by a transcript encoder architecture, where $L_T$ is the length of the encoded transcript representation (typically equal to the length of the input transcript) and $d_T$ is the embedding dimension produced by the transcript encoder. For each speaker in the dataset, a $\Re^{d_S}$ embedding vector is initialized with Glorot \cite{glorot2010understanding} initialization. For each example, the $d_S$-dimensional speaker embedding corresponding to the true speaker of the example is broadcast-concatenated with the $L_T \times d_T$-dimensional transcript encoder representation to form a $(d_T + d_S)$-dimensional sequence of encoder embeddings that the decoder will attend to. No additional changes or loss metrics are necessary.

\subsection{Reference Encoder}
\label{sec:architecture-encoder}

We extend the Tacotron architecture by adding a ``reference encoder'' module that takes a length-$L_R$ and $d_R$-dimensional reference signal as input, and computes a $d_P$-dimensional embedding from it. We think of this fixed-dimensional embedding as the ``prosody space'' -- our goal is that sampling from this space will yield diverse and plausible output speech, and that we can manipulate elements of this space to control the output meaningfully.

As with the speaker embedding, this prosody embedding is combined with the $L_T \times d_T$ text encoder representation via a broadcast-concatenation. In combination with the speaker embeddings described in \S\ref{sec:multi-speaker}, the encoder embeddings form a $L_T \times (d_T + d_S + d_P)$ embedding matrix, where the speaker and prosody embeddings are fixed across all timesteps. Figure \ref{fig:tacotron-conditioning} illustrates this structure.

During training, the reference acoustic signal is simply the target audio sequence being modeled. No explicit supervision signal is used to train the reference encoder; it is learned using Tacotron's reconstruction error as its only loss. In training, one can think of the combined system as an RNN encoder-decoder \cite{cho2014learning} with phonetic and speaker information as conditioning input. For a sufficiently high-capacity embedding, this representation could simply learn to copy the input to the output during training. Therefore, as with an autoencoder, care must be taken to choose an architecture that sufficiently bottlenecks the prosody embedding such that it is forced to learn a compact representation.

During inference, we can use the prosody reference encoder to encode \emph{any} utterance: we are not constrained to match either the text input or the provided speaker embedding. In particular, this enables the possibility of prosody transfer -- using an utterance by a different speaker, or different text to control the output. We study prosody transfer in detail in \S\ref{sec:experiments}.

For the reference encoder architecture (Figure \ref{fig:reference-encoder}), we use a simple 6-layer convolutional network. Each layer is composed of $3 \times 3$ filters with $2 \times 2$ stride, SAME padding and ReLU activation. Batch normalization \cite{ioffe2015batch} is applied to every layer. The number of filters in each layer doubles at half the rate of downsampling: 32, 32, 64, 64, 128, 128.

The $L_R \times d_R$ reference signal is downsampled by this architecture $64$ times in both dimensions. The $\lceil d_R / 64 \rceil$ feature dimensions and $128$ channels of the final convolution layer are unrolled as the inner dimension of the resulting $\lceil L_R / 64 \rceil \times 128 \lceil d_R/64 \rceil$ matrix. To compress the $\lceil L_R / 64 \rceil$-length sequence produced by the CNN layers down to a single fixed-length vector, we use a recurrent neural network with a single 128-width Gated Recurrent Unit (GRU) \cite{cho2014gru} layer. We take the final 128-dimensional output of the GRU as the pooled summarization of the sequence.

To compute the final $d_P$-dimensional embedding from the 128-dimensional output of the GRU, we apply a fully-connected layer to project the output to the desired dimensionality, followed by an activation function (e.g. softmax, tanh). The choice of activation function can constrain the information contained in the embedding and make learning easier by controlling its magnitude. After some exploration, we found that a $d_P$ of 128 and a tanh activation perform well in practice.

\subsection{Reference signal feature representation}
\label{sec:architecture-feature}

The choice of $L_R \times d_R$ feature representation used as the input to the reference encoder architecture naturally impacts the aspects of prosody we can expect to learn. For example, a pitch track representation will not allow us to model prominence in some languages since it does not contain energy information. Similarly an MFCC representation may be somewhat pitch-invariant (depending on the number of coefficients retained), preventing us from modeling intonation. In this work, we decided to use the same perceptually-relevant summarization of the spectrum that \cite{wang2017} does: the mel-warped spectrum \cite{stevens1937scale}.

This choice of representation enables an interpretation of the resulting architecture as an RNN encoder-decoder \cite{cho2014learning} conditioned on text and speaker identity. All it must model via its bottleneck representation is the unexplained variation in the signal, i.e. the prosody and recording environment. We illustrate this interpretation in Figure \ref{fig:tacotron-autoencoder}.

We explored more compact representations, such as pitch track and intensity, instead of mel spectrograms, which also produced successful results, but we focus on mel spectrograms in this paper.

\subsection{Variable-Length Prosody Embeddings}
\label{sec:architecture-variable}

The use of a fixed-length prosody embedding poses an obvious scaling bottleneck, preventing the extension of this approach to longer utterances.
An alternate implementation of the reference encoder in \S\ref{sec:architecture-encoder} uses the output of the GRU at every time step rather than just the final output.  As with the fixed-length encoder, each GRU output is passed through a fully connected layer to transform it to the desired dimensionality.
This can be interpreted as a low-bitrate representation of prosody similar to the proposal of Enhanced Phonetic Transcriptions in \cite{coile1994protran}. 
To condition the Tacotron decoder on this sequence, we introduce a second attention head with an attention-aggregator module as proposed in \cite{wang2017uncovering}. 

In our experiments, variable-length prosody embeddings are able to generalize to very long utterances; however, compared to fixed-length embeddings, variable-length embeddings are not as robust to text and speaker perturbations likely because they encode a stronger timing signal. 
Therefore, this paper focuses on fixed-length embeddings.
\section{Experiments and Results}
\label{sec:experiments}

\begin{table*}[t]
\caption{A summary of quantitative and subjective metrics (\S\ref{sec:experiments-metrics}) used to evaluate the prosody transfer. Lower is better for both MCD$_k$ and FFE. Higher subjective scores are better, and indicate whether human raters believe the voice is closer in prosody to the reference than the corresponding baseline model on a 7 point ($-3$ to $3$) scale, where 0 is ``about the same''.}
\label{table:same-text-prosody-transfer}
\vskip 0.15in
\begin{center}
\begin{small}
\begin{sc}
\begin{tabular}{lcccccr}
\toprule
Voice & Model & Reference               & MCD$_{13}$ & FFE & Subjective \\
\midrule
Single-speaker & baseline & same speaker   & 10.63 & 53.2\% & \\
Single-speaker & tanh-128 & same speaker   & $\bm{7.92}$ & $\bm{28.1\%}$ & $\bm{1.611 \pm 0.164}$ \\
\midrule
Single-speaker & baseline & unseen speaker & 11.22 & 59.6\% & \\
Single-speaker & tanh-128 & unseen speaker & $\bm{8.89}$ & $\bm{38.0\%}$ & $\bm{1.465 \pm 0.132}$ \\
\toprule
Multi-speaker & baseline & same speaker   & 9.93 & 48.5\% & \\ 
Multi-speaker & tanh-128 & same speaker   & $\bm{6.99}$ & $\bm{27.5\%}$ & $\bm{1.307 \pm 0.127}$ \\
\midrule
Multi-speaker & baseline & seen speaker   & 12.37 & 64.2\% & \\
Multi-speaker & tanh-128 & seen speaker   & $\bm{9.51}$ & $\bm{37.1\%}$ & $\bm{0.871 \pm 0.138}$ \\
\midrule
Multi-speaker & baseline & unseen speaker & 11.84 & 60.0\% & \\
Multi-speaker & tanh-128 & unseen speaker & $\bm{10.87}$ & $\bm{41.3\%}$ & $\bm{1.146 \pm 0.246}$ \\
\bottomrule
\end{tabular}
\end{sc}
\end{small}
\end{center}
\vskip -0.1in
\end{table*}

\subsection{Datasets and training}
We use the following datasets:
\begin{description}
    \item[Single-speaker dataset:] A single speaker high-quality English dataset of audiobook recordings by Catherine Byers (the speaker from the 2013 Blizzard Challenge). This dataset consists of 147 hours of recordings of 49 books, read in an animated and emotive storytelling style.
    \item[Multi-speaker dataset:] A proprietary high-quality English speech dataset consisting of 296 hours across 44 speakers (5 with Australian accents, 6 with British accents, 1 with an Indian accent, 2 with Singaporean accents, and 30 with United States accents).
\end{description}
We train our models for at least 200k steps with a minibatch size of 256 using the Adam optimizer \cite{kingma2014adam}. We start with a learning rate of \num{1e-3} and decay it to \num{5e-4}, \num{3e-4}, \num{1e-4}, and \num{5e-5} at step 50k, 100k, 150k, and 200k respectively. For baselines, we train models without the reference encoder architecture (\S\ref{sec:architecture}).

\subsection{Evaluation metrics}
\label{sec:experiments-metrics}

There are no generally-accepted metrics for prosody transfer. To measure performance, we adapt a number of metrics from general audio processing, each of which reflects an acoustic correlate of prosody. For all comparisons of predicted signals to target signals, we extend the shorter signal to the length of the longer signal using a domain-appropriate padding (e.g. $0$ for a time-domain waveform, $-13.8$ for a log magnitude spectrogram with a \num{1e-6} stabilizing offset). All pitch and voicing metrics are computed using the output of the YIN \cite{de2002yin} pitch tracking algorithm. 

\begin{description}
\item[Mel Cepstral Distortion (MCD$_K$) \cite{kubichek1993mel}:]

\begin{align*}
\textrm{MCD}_K = \frac{1}{T}\sum_{t=0}^{T-1}\sqrt{\sum_{k=1}^K \left(c_{t,k} - c_{t,k}'\right)^2}
\end{align*}
Where $c_{t,k}$,$c_{t,k}'$ are the $k$-th mel frequency cepstral coefficient (MFCC) of the $t$-th frame from the reference and predicted audio. We sum the squared differences over the first $K$ MFCCs, skipping $c_{t,0}$ (overall energy).

\item[Gross Pitch Error (GPE) \cite{nakatani2008method}:] 
\begin{align*}
\textrm{GPE} = \frac{\sum_t \mathbbm{1}\left[\left|p_t - p_t'\right| > 0.2 p_t\right] \mathbbm{1}[v_t]\mathbbm{1}[v_t']}{\sum_t \mathbbm{1}[v_t]\mathbbm{1}[v_t']}
\end{align*}

Where $p_t$,$p_t'$ are the pitch signals from the reference and predicted audio, $v_t$,$v_t'$ are the voicing decisions from the reference and predicted audio, and $\mathbbm{1}$ is the indicator function. 
The GPE measures the percentage of voiced frames that deviate in pitch by more than 20\% compared to the reference.

\item[Voicing Decision Error (VDE) \cite{nakatani2008method}:] 
\begin{align*}
\textrm{VDE} = \frac{\sum_{t=0}^{T-1} \mathbbm{1}[v_t \neq v_t']}{T}
\end{align*}
Where $v_t$,$v_t'$ are the voicing decisions for the reference and predicted audio, $T$ is the total number of frames, and $\mathbbm{1}$ is the indicator function.

\item[F0 Frame Error (FFE) \cite{chu2009reducing}:] 
\begin{align*}
\frac{\sum_{t=0}^{T-1} \mathbbm{1}\left[\left|p_t - p_t'\right| > 0.2 p_t\right] \mathbbm{1}[v_t]\mathbbm{1}[v_t'] + \mathbbm{1}[v_t \neq v_t']}{T}
\end{align*}

FFE measures the percentage of frames that either contain a 20\% pitch error (according to GPE) or a voicing decision error (according to VDE).

\end{description}

In addition to these metrics, we propose a subjective (i.e., human) test structured as an AXY discrimination test that we refer to as an ``anchored prosody side-by-side''. A human rater is presented with three stimuli: a reference speech sample (A), and two competing samples (X and Y) to evaluate. The rater is asked to rate whether the prosody of X or Y is closer to that of the reference on a 7-point scale. The scale ranges from ``X is much closer'' to ``Both are about the same distance'' to ``Y is much closer'', and can naturally be mapped on the integers from $-3$ to 3. Prior to collecting any ratings, we provide the raters with 4 examples of prosodic attributes to evaluate (intonation, stress, speaking rate, and pauses), and explicitly instruct the raters to ignore audio quality or pronunciation differences. A screenshot of this user interface is included in Figure \ref{fig:subjective-evaluation-template}. For each triplet (A, X, Y) evaluated, we collect 4 independent ratings. No rater is used for more than 6 items in a single evaluation. To analyze the data from these subjective tests, we average the scores and compute 95\% confidence intervals.

\subsection{Same-text Prosody Transfer}

We first demonstrate that our model is capable of prosody transfer when the text is unchanged from that of the reference utterance. 

\subsubsection{Spectrograms and Pitch Tracks}

\begin{figure}[tb]
\vskip 0.2in
\begin{center}
\centerline{\includegraphics[width=\columnwidth]{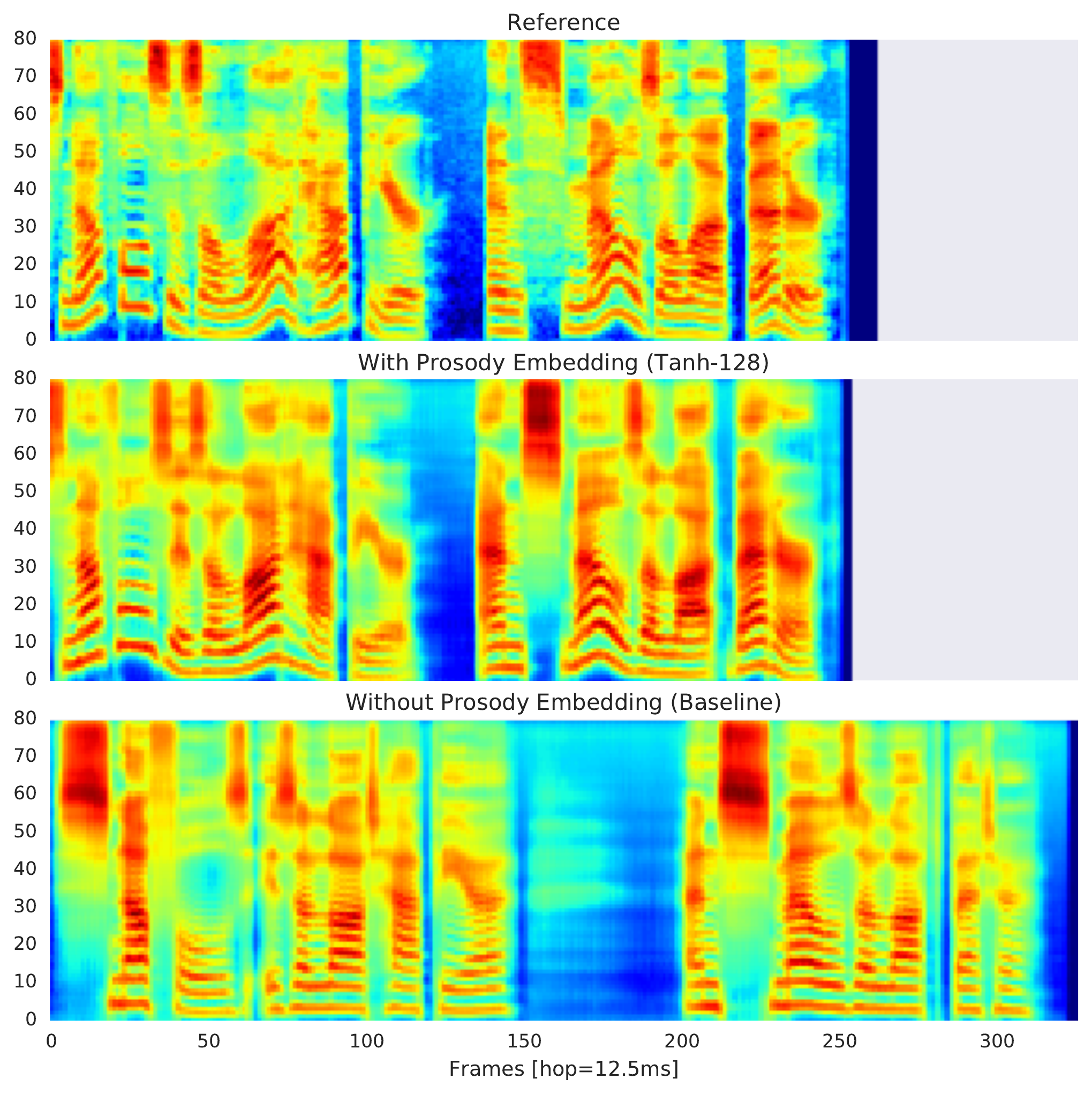}}
\caption{Mel spectrograms for the utterance ``Snuffles is a lot happier. And smells a lot better.''  (Top) Reference utterance from an unseen speaker.  (Middle) Synthesized utterance conditioned on reference embedding.  (Bottom) Synthesized utterance from a model without reference conditioning.}
\label{fig:specgrams}
\end{center}
\vskip -0.2in
\end{figure}

Figure~\ref{fig:specgrams} shows three spectrograms (reference, baseline model, prosody embedding model) for the same utterance. Note that the spectrogram from the model conditioned on a reference embedding bears a much stronger resemblance to the reference signal than that generated by an unconditioned model. In particular, notice that the spectrogram from the baseline model, which does not use a reference signal, exhibits noticeably different rhythm -- for example, there is a long pause between the two halves of the utterance, and the utterance lasts much longer. By contrast, the output with a prosody embedding has the same length and pause characteristics as the reference audio; it also has recognizably similar harmonic and onset structure.

\begin{figure}[tb]
\vskip 0.2in
\begin{center}
\centerline{\includegraphics[width=\columnwidth]{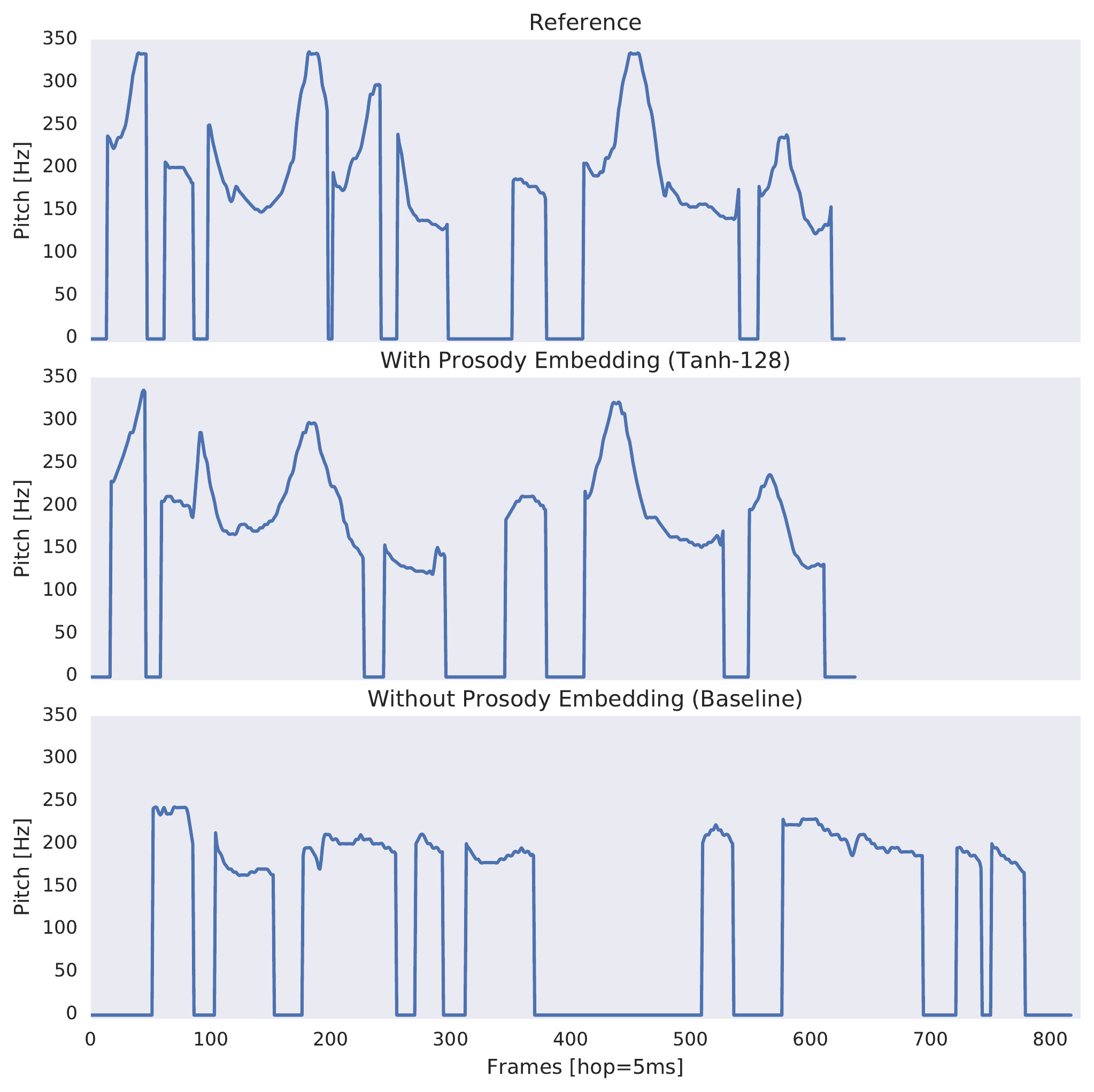}}
\caption{Pitch tracks for the utterance ``Snuffles is a lot happier. And smells a lot better.''  A pitch of 0~Hz indicates an unvoiced segment. (Top) Reference utterance from an unseen speaker.  (Middle) Synthesized utterance conditioned on reference embedding.  (Bottom) Synthesized utterance from a model without reference conditioning.}
\label{fig:pitch_tracks}
\end{center}
\vskip -0.2in
\end{figure}

Figure~\ref{fig:pitch_tracks} shows the pitch tracks for the same triplet of utterances. We can see that the prosody embedding model closely follows the pitch contours of the reference, whereas the unconditioned model does something else entirely.

\subsubsection{Quantitative and Subjective Evaluations}

We evaluated synthesis of single- and multi-speaker models using two types of reference utterance. ``Same speaker'' indicates a reference utterance from the same speaker as the target, while ``unseen speaker'' refers to a reference utterance from a speaker unseen in training. For the multi-speaker model, we also tested synthesis with a speaker seen in training but different from the target speaker (``seen speaker''). 

We present our findings in Table \ref{table:same-text-prosody-transfer}. The results show that augmenting Tacotron with a reference encoder allows it to match the reference prosody substantially more accurately. This is true for all baseline/model pairs in Table \ref{table:same-text-prosody-transfer}, and is independent of whether the reference speaker matches the target speaker. The objective metrics MCD$_{13}$ and FFE also support this conclusion, both resulting in substantially lower values for the reference encoder model than for the baseline model.

Note that when the target and reference speakers are different (i.e., when the reference in Table \ref{table:same-text-prosody-transfer} is either ``seen speaker'' or ``unseen speaker''), we must be careful to demonstrate that prosody transfer has been achieved. 
If the bottleneck allows too much information to flow through the reference encoder, for example, the overall model could simply copy the reference to the output. In this instance, listening to even a small number of outputs suffices to verify that the output speaker matches the target speaker, and that we have in fact achieved prosody transfer across speakers. However, further experiments, explored in \S\ref{sec:speaker-id}, provide some surprising results.

\subsection{Templated Prosody Transfer}\label{sec:templated-prosody-transfer}

In addition to same-text prosody transfer, we also explore the robustness of our proposed model to changes in the synthesized text. Since the prosody embeddings we learn capture prosodic features with some fine time detail, it isn't clear what it would mean to transfer these prosodic features to a radically different utterance. As expected, we find that drastic changes to the sentence or phrase structure result in undesirable prosody transfer. This use case may be more suited to models that capture less granular features of prosody such as emotion or style. \cite{wang2018}, for example, applies a similar approach to learning representations of global style.

Nonetheless, we include a number of examples on our \demopagelink{demo page} demonstrating that text transformations can be performed without compromising intelligibility or desired prosody. This can be highly useful in building templated dialogue systems capable of synthesizing a template with a desired prosody.

\subsection{Preservation of Speaker Identity}\label{sec:speaker-id}
In Table \ref{table:same-text-prosody-transfer}, the results of our ``anchored prosody side-by-side'' subjective evaluation show that reference-based synthesis matches the reference audio significantly better than the baseline model. However, the evaluation does not assess whether the target speaker identity was preserved by the synthesis. This is not accidental: pitch, pacing, and other prosodic characteristics factor into speaker identity, and thus it is difficult to prescribe exactly which aspects of the target speaker's identity should be preserved during prosody transfer. 

The audio samples we include on our \demopagelink{demo page} show that our model preserves many important aspects of speaker identity during prosody transfer. We include a grid of audio examples representative of typical performance of this system, with reference clips from 6 speakers with distinct accents. Each utterance is synthesized 6 times, each with a different target speaker. Notably, the prosody of each clip matches that of the reference, while the distinct accents and vocal tract properties of each speaker are preserved.

However, listening to samples of a male voice controlling a female voice (and vice-versa) reveals that our prosody representation encodes pitch in an absolute manner. When controlled by a male reference signal, female target speakers sound as if they're imitating a person with a deeper voice. Similarly, when controlled by a female reference signal, male speakers sound as if they're imitating a person with a higher voice. This suggests that the prosody and speaker representations are somewhat entangled.

To quantify this entanglement, we designed a simple speaker identification model that takes varying types of acoustic input, and produces predictions of speaker identity from a universe of speakers known at training time. The architecture uses the same strided convolutions and GRU-based aggregation as the reference encoder architecture from \S\ref{sec:architecture-encoder}, and is independently trained on ground truth mel spectrograms using the same 44-speaker dataset used to train our multi-speaker model. The architecture achieves over 99\% accuracy on both the held-out ground truth and synthesized audio from our baseline 44-speaker model.

We then tested our prosody-enhanced Tacotron using this model. To do so, we first constructed pairs of all target speakers and reference utterances in the test set. We then used our prosody-enhanced Tacotron to generate mel spectrograms for these pairs, and fed the output into the speaker identification model. The speaker identification model identified the spectrograms as originating from the reference speaker in 61\% of test set examples, and the target speaker only 21\% of the time (ideally, the target speaker would be at 100\%). We refer the reader to the \demopagelink{audio samples} to understand how surprising this is -- the audio samples \emph{sound} substantially more like the target speaker in every sample we've listened to. 

Since our model seems to transfer prosody in a pitch-absolute manner, we ran a further experiment where we trained the speaker identification model on 13 Mel-frequency cepstral coefficients (MFCCs) which contain less pitch content. In this case, the speaker identification model identified the utterances as originating from the reference speaker 41\% of the time, and the target speaker 32\% of the time, suggesting that, indeed, speaker-dependent pitch content is transferred from the reference to the output.

\subsection{Bottleneck Size and Shape}
\label{sec:experiments-bottleneck}

\begin{figure}[htb]
\vskip 0.2in
\begin{center}
\centerline{\includegraphics[width=\columnwidth]{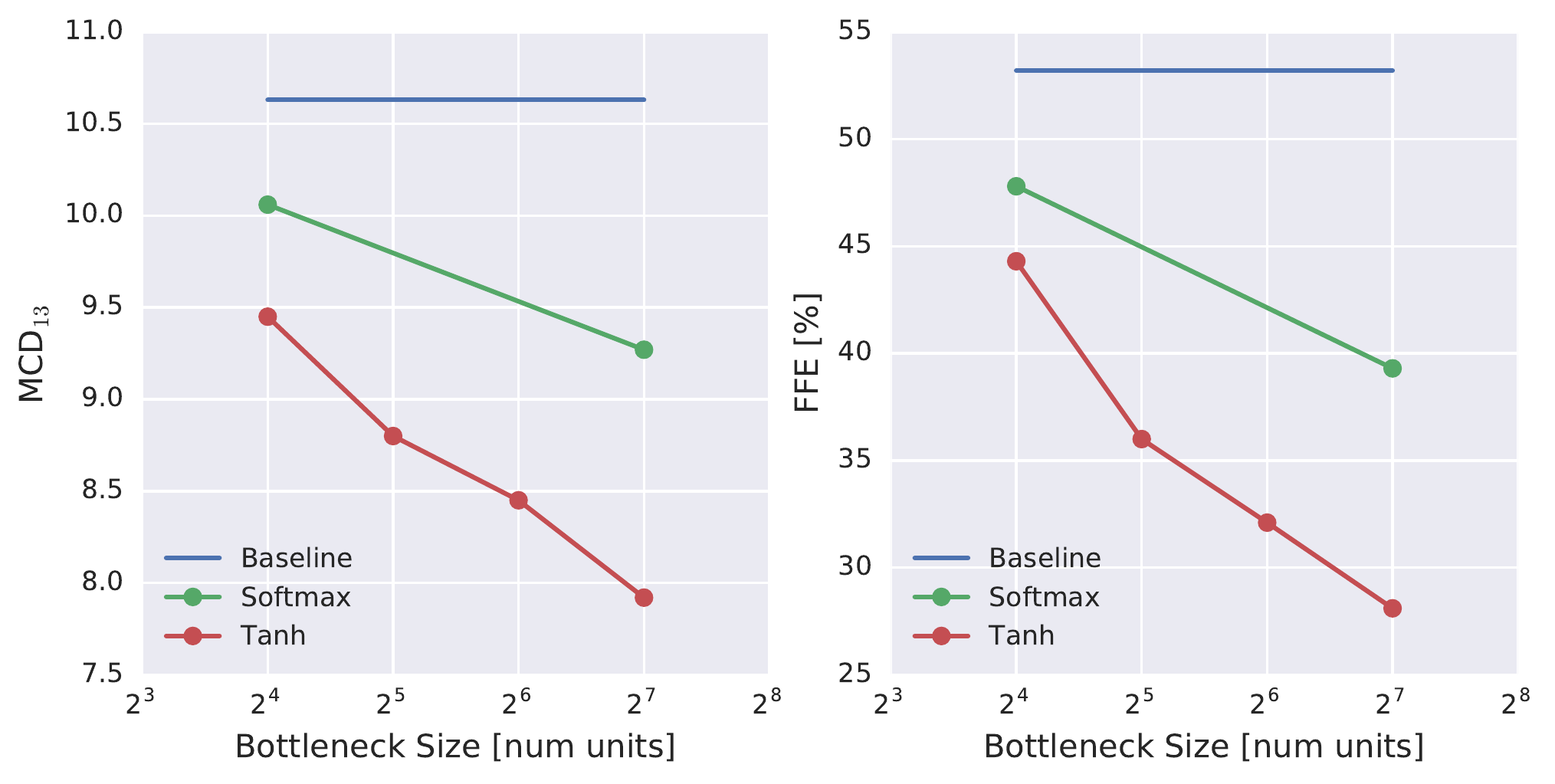}}
\caption{The effect of bottleneck size on quantitative metrics. In terms of both MCD$_{13}$ and FFE, models with prosody encoders beat the baseline. As the bottleneck size increases, the performance in both metrics improve. Softmax is a more severe bottleneck than tanh, and exhibits worse metrics.}
\label{fig:bottleneck-plots}
\end{center}
\vskip -0.2in
\end{figure}

The dimensionality and activation used for the bottleneck substantially affect the information flow from the prosody reference encoder to the output. In this experiment, we use our single speaker as both the reference signal and target (we are essentially trying to conditionally autoencode the mel spectrograms given text). We plot the MCD$_{13}$ and FFE metrics while varying the bottleneck size and activation in Figure \ref{fig:bottleneck-plots}, and include a series of audio samples on our \demopagelink{demo page}. We can conclude that increasing the bottleneck size allows for significantly more information flow from the reference to the output, allowing for better reproduction of the reference. More interestingly, using a softmax activation leads to a degradation of metrics in comparison to tanh: this is probably due to the exponential suppression of the non-maximal components in the softmax.

The quantitative metrics are in agreement with the audio samples provided on our \demopagelink{demo page}: larger bottlenecks with the tanh activation improve audio similarity, and the outputs are more faithful to the reference prosody. 
A potential trade-off is that a narrower bottleneck would likely better preserve the speaker identity of the target speaker.
\section{Discussion and Future Work}
\label{conclusion}

In this work, we have demonstrated prosody transfer via an end-to-end learned representation of prosody directly from acoustic signals.
While our system successfully transfers prosody from one speaker to another, it does so in a pitch-absolute manner.
Future work should focus on encoding prosody in a pitch-relative manner so that speaker identity is more completely preserved during transfer.  

A substantial open question is how to disentangle the textual information implicit in the reference signal from the prosodic information. In Section \ref{sec:templated-prosody-transfer}, we showed that this is possible to some extent, especially when the transcripts are relatively close. But, more generally, this amounts to transferring or controlling prosody using utterances with different corresponding text transcripts. As noted earlier, this is a somewhat ill-defined task, and a more careful formalization of this problem is needed to make real progress.

We also defined objective and subjective metrics for evaluating prosody transfer, and evaluated our architecture on these benchmarks.  
Solidifying metrics that quantify all desired aspects of prosody transfer (e.g., prosodic similarity and the degree to which prosodic, textual, and speaker information are disentangled) is an important step in the long-term progression of end-to-end prosody work.

Finally, given our construction of a prosody space, we would like to be able to sample from this space (i.e., generate prosody instead of transferring it). One could, for example, attempt to learn a prior distribution over the prosody space.

\section*{Acknowledgements}

The authors thank Aren Jansen and the Machine Hearing, Google Brain and Google TTS teams for their helpful discussions and feedback.

\bibliography{bibliography}
\bibliographystyle{icml2018}

\onecolumn
\appendix 

\section{Subjective Evaluation Template}
\label{sec:subjective-evaluation-template}

\begin{figure}[h]
\vskip 0.2in
\begin{center}
\centerline{\includegraphics[width=\textwidth]{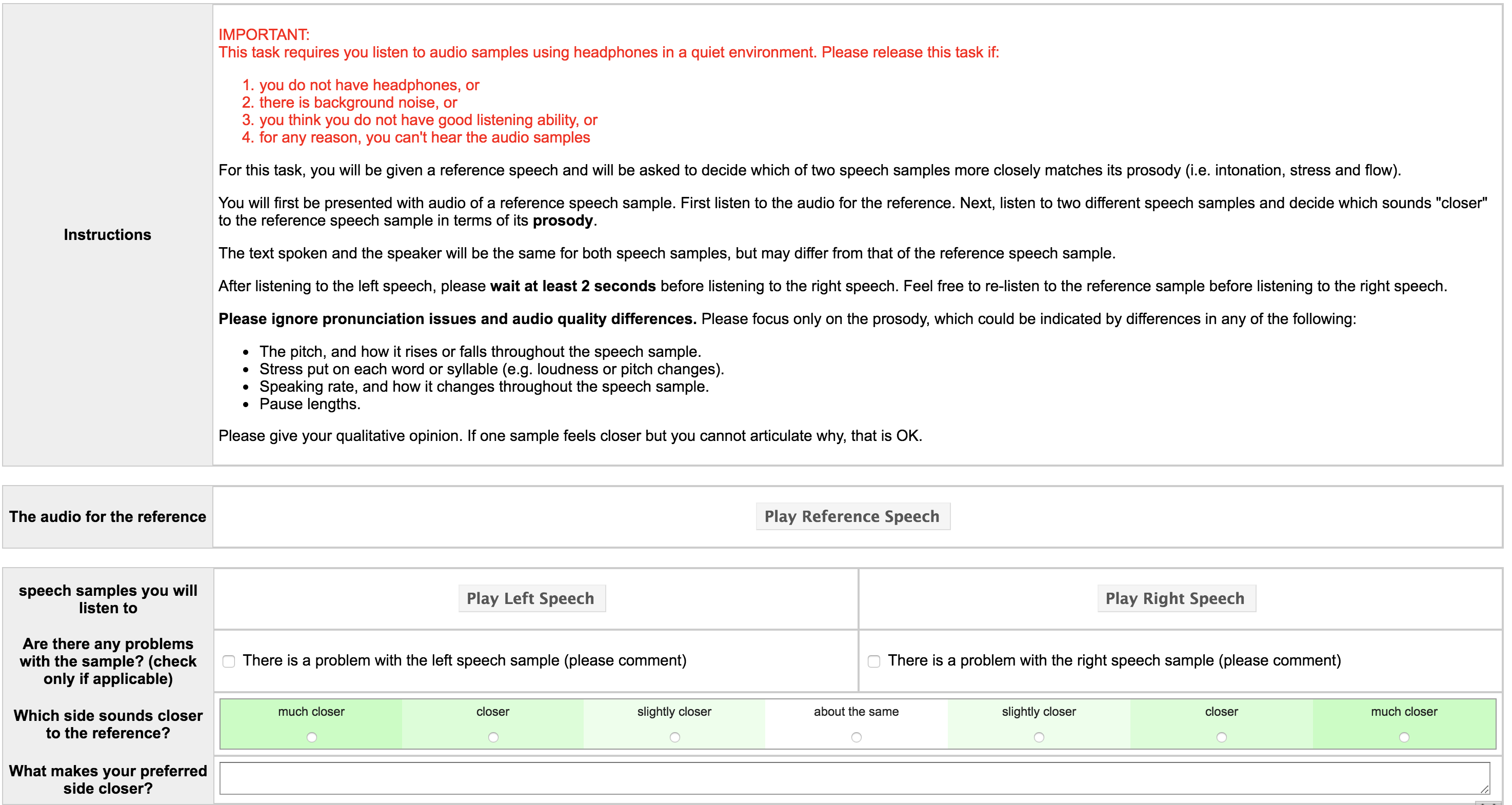}}
\caption{The subjective evaluation template described in Section \ref{sec:experiments-metrics}.
A human rater is presented with three stimuli: a reference speech sample (A), and two competing samples (X and Y) to evaluate. The rater is asked to rate whether the prosody of X or Y is closer to that of the reference on a 7-point scale. The scale ranges from ``X is much closer'' to ``Both are about the same distance'' to ``Y is much closer'', and can naturally be mapped on the integers from $-3$ to 3. Prior to collecting any ratings, we provide the raters with 4 examples of prosodic attributes to evaluate (intonation, stress, speaking rate, and pauses), and explicitly instruct the raters to ignore audio quality or pronunciation differences. For each triplet (A, X, Y) evaluated, we collect 4 independent ratings. No rater is used for more than 6 items in a single evaluation. To analyze the data from these subjective tests, we average the scores and compute 95\% confidence intervals.}
\label{fig:subjective-evaluation-template}
\end{center}
\vskip -0.2in
\end{figure}

\end{document}